# Capsule GAN Using Capsule Network for Generator Architecture

Kanako Marusaki and Hiroshi Watanabe
Graduate School of Fundamental Science and Engineering, Waseda University
Tokyo, Japan
Email: k-marusaki@suou.waseda.jp

*Abstract*— This paper presents Capsule GAN, a Generative adversarial network using Capsule Network not only in the discriminator but also in the generator. Recently, Generative adversarial networks (GANs) has been intensively studied. However, generating images by GANs is difficult. Therefore, GANs sometimes generate poor quality images. These GANs use convolutional neural networks (CNNs). However, CNNs have the defect that the relational information between features of the image may be lost. Capsule Network, proposed by Hinton in 2017, overcomes the defect of CNNs. Capsule GAN reported previously uses Capsule Network in the discriminator. However, instead of using Capsule Network, Capsule GAN reported in previous studies uses CNNs in generator architecture like DCGAN. This paper introduces two approaches to use Capsule Network in the generator. One is to use DigitCaps layer from the discriminator as the input to the generator. DigitCaps layer is the output layer of Capsule Network. It has the features of the input images of the discriminator. The other is to use the reverse operation of recognition process in Capsule Network in the generator. We compare Capsule GAN proposed in this paper with conventional GAN using CNN and Capsule GAN which uses Capsule Network in the discriminator only. The datasets are MNIST, Fashion-MNIST and color images. We show that Capsule GAN outperforms the GAN using CNN and the GAN using Capsule Network in the discriminator only. The architecture of Capsule GAN proposed in this paper is a basic architecture using Capsule Network. Therefore, we can apply the existing improvement techniques for GANs to Capsule GAN.

## I. INTRODUCTION

Recently, image processing using convolutional neural networks (CNNs) [1] has been intensively studied. Generative adversarial networks (GANs) [2], which generate artificial images, is one of them. There are a lot of GANs that use CNNs such as Deep Convolutional GAN (DCGAN) [3]. However, GANs sometimes fail to produce realistic images. This is because, the generator of GANs tends to collapse, and GANs are greatly influenced by the hyperparameter settings, making it difficult to balance learning between the discriminator and the generator. Therefore, learning is not stable because of the parameters tend to drop in the local optimum.

CNNs have a pooling layer. While a pooling layer reduces the size of the data and makes it easier to handle to process, CNNs loses information about the relationships between image features [4]. Therefore, CNNs tend to recognize an object when the parts of the object are complete, even if the position of the parts changes. Hinton et al. [4] proposed Capsule Network in 2017. Capsule Network overcomes such defect of CNNs. Thus, we expect that GANs using Capsule Network is better than GAN using CNNs to preserve the position of objects in the image. GANs (Capsule GAN1 in Fig.1) using Capsule Network are proposed [5], [6]. However, these GANs use Capsule Network only in the discriminator, not in the generator.

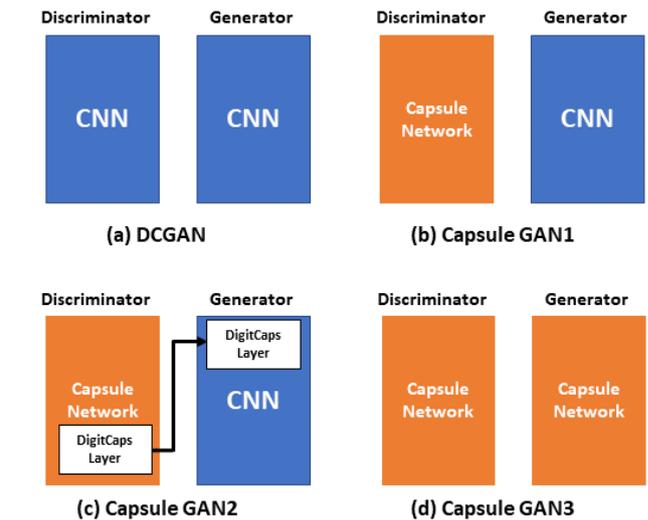

Fig.1. The structures of GANs used in this experiment. (a) is DCGAN, using CNN. (b), (c) and (d) are GAN using Capsule Network. (b) is proposed by Gadirov et al. [5]. This structure uses Capsule Network in the discriminator only. (c) and (d) use Capsule Network in the discriminator and the generator.

Therefore, we propose Capsule GAN that uses Capsule Network in both the discriminator and the generator architecture of GAN. We introduce two approaches to using Capsule Network in the generator. One is to use the DigitCaps layer of the discriminator for the generator input [7] (Capsule GAN2 in Fig.1). DigitCaps layer is the output layer of Capsule Network. The other is to use Capsule Network in the generator such as CNNs (Capsule GAN3 in Fig.1). We performed an experiment to generate images by Capsule GANs and to evaluate the images generated by Capsule GANs. In these experiments, we use MNIST [8], Fashion-MNIST [9] and "cat images" as training images. The cat images are color images, which is the dataset [10] published by Oxford university. We use Inception Score [11] for evaluation of the generated images by GANs.

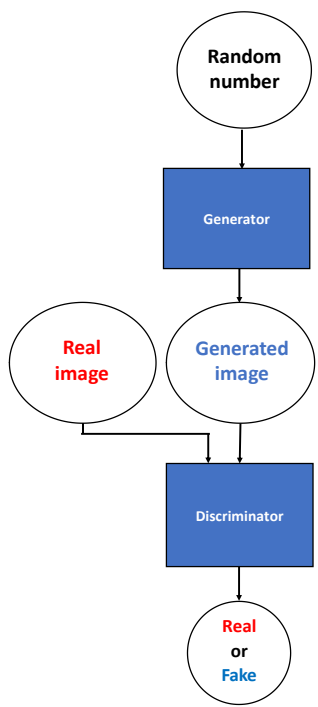

Fig.2. The structure of GANs

Capsule GAN architecture is a simple GAN architecture. Therefore, GAN's improvement methods [14], [15] are applicable to Capsule GAN. The purpose of this study is to improve the quality of the images generated by Capsule GAN.

## II. RELATED WORK

### A. Generative Adversarial Networks Using CNN

GANs using CNNs have been intensively investigated. DCGAN is one of the typical examples that tries to improve generated image quality with CNN. There are also Conditional GANs [12] and Auxiliary Classifier GANs [13] with labels indicating the type of images generated. In addition, there are Wasserstein GAN [14] and Wasserstein GAN-gp [15] that use Wasserstein distance to prevent mode collapse of GANs. Moreover, Big GAN [16] and Style-based GAN [17] generates high resolution images. However, learning of GANs is not stable because of the parameters tend to drop in the local optimum. Thus, there are still problems such as mode collapse [18].

### B. Generatice Adversarial Networks Using Capsule Network

Some GANs using Capsule Network are reported. Gadirov et al. [5] and Jaiswal et al. [6] proposed Capsule GAN (Capsule GAN1). Capsule GAN1 improved the generation of MNIST and CIFAR-10 [19] images over DCGAN, a simple GAN using CNN in [6]. Capsule GAN1 uses Capsule Network in the discriminator. However, instead of using Capsule Network, Capsule GAN1 uses CNNs in generator architecture such as DCGAN. We reported Capsule GAN using Capsule Network in the generator [7] (Capsule GAN2), using the Capsule Network layer of the discriminator for the generator input. However, this Capsule GAN generated only MNIST images in [7]. In this paper, we show the difficulty to generate realistic images by the conventional Capsule GAN through the additional experiment with more complicated images such as Fashion-MNIST and "cat images".

In this paper, we propose a GAN architecture using Capsule Network not only in the discriminator but also in the generator.

## III. GENERATIVE ADVERSARIAL NETWORKS

GANs are models that generate images similar to training data. GANs are composed of two networks, the discriminator and the generator. Fig. 2 shows the structure of GANs. The generator outputs images like training images from random numbers. The inputs to the discriminator are the images generated by the generator and the training images. The discriminator distinguishes whether the input image is the training image (real) or the generated image (fake). The generator is trained to be able to deceive the discriminator, and the discriminator is trained not to be deceived by the generator. In other words, the discriminator and the generator play an adversarial game. We defined random number distribution as $z \sim p_z$, the training data distribution as $x \sim p_x$, generated data as $G(z)$ and the output of the discriminator as $D(\cdot)$. The equation of loss function is given by

$$\min_G \max_D V(D,G) = \mathbb{E}_{x \sim p_{data}(x)}[logD(x)] + \mathbb{E}_{z \sim p_z(z)}[log(1-D(G(z)))] \quad (1)$$

They use the Wasserstein distance in the calculation of the loss function to prevent from mode collapse in [14], [15]. It does not change the GAN architecture. Therefore, these methods will be applicable in the future to Capsule GAN proposed in this paper.

## IV. CAPSULE NETWORK

Hinton et al. [4] proposed Capsule Network in 2017. Capsule Network is a neural network based CNNs. CNNs have the defect that the relational information between features of the image may be lost. Capsule Network overcomes the defect in two ways. One is to process layers without pooling. The other is that the input of neurons is a vector, not a scalar. This vector is called a capsule. Fig. 3 shows the architecture of Capsule Network. Input vector $u_i$ in the layer $l_i$ is multiplied by weight matrix $W$. The magnitude of $u_i$ indicates the existence probability of the objects. The detection of $u_i$ indicates the relational information of the objects. The equation is given by

$$\hat{u}_{j|i} = W_{ij} u_i \quad (2)$$

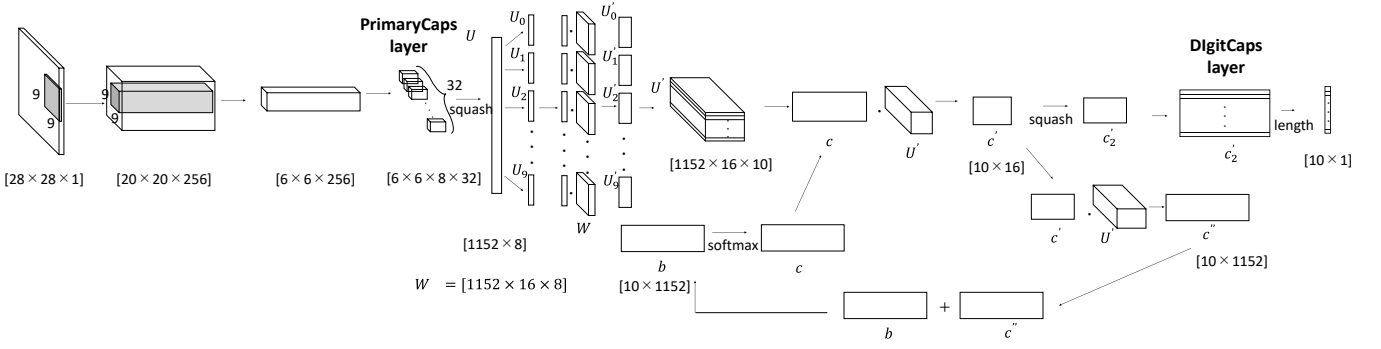

Fig.3. The architecture of Capsule Network. This architecture is for 28×28 pixel images, MNIST. First, input images are calculated by convolution layers. There is no pooling layer in this architecture. Second, the vectors (capsules) are prepared for the number of classification class (10 classes). Each capsule is multiplied by weight matrix. This process is called dynamic routing. After dynamic routing, DigitCaps layer is outputted. DigitCaps layer has 16 dimensions vectors that have features of the image. The length of vectors represents the probability of the classification.

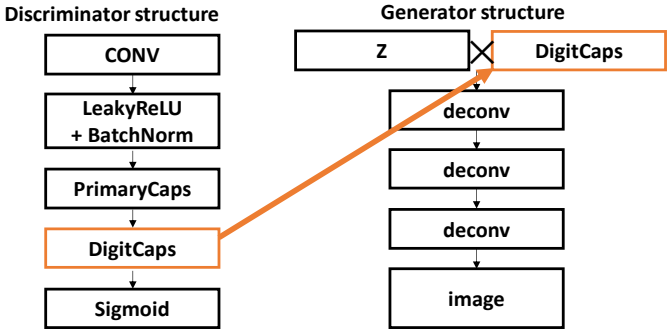

Fig.4. The structure of Capsule GAN2 using DigitCaps layer in the generator. DigitCaps layer is multiplied by latent values. The calculated values are inputs of the generator.

In Capsule Network, weights are trained by a routing called dynamic routing. The weight is defined as $c_{ij}$, a capsule in layer $l_i$ is defined as $i$, a capsule in layer $l_j$ is and capsule $j$ is defined as $j$. $c_{ij}$ is given by

$$c_{ij} = \frac{exp(b_{ij})}{\sum_k exp(b_{jk})} \quad (3)$$

$b_{ij}$ is defined as 0 at first. $b_{ij}$ is given by

$$b_{ij} \leftarrow b_{ij} + \hat{u}_{j|i} \cdot v_j \quad (4)$$

Where $v$ is the output vector. The above process is repeated. As a result, equation (4) is increased if it is an important value. The output layer after dynamic routing is called DigitCaps layer.

Capsule Network uses the squashing function as activation function. The input vector is defined as $s$. The output vector $v$ is defined as follows

$$v = \frac{\|s\|^2}{1+\|s\|^2} \frac{s}{\|s\|} \quad (5)$$

The squashing function scales the magnitude to 1 without changing the orientation of the vector.

V. INCEPTION SCORE

Inception Score (IS) [8] is a measure of evaluating the quality of images generated by GANs. If the image can be easily identified with the Inception model [20] and the diversity of identified labels increases, the Inception Score will increase. It evaluates that the large the value of Inception Score, the better the image. We define $i$-th image as $X_i$, the label as $y$, the probability of label $y$ inputting $X_i$ to the Inception model as $p(y|x_i)$, the probability of label $y$ of all images and the set of images as $X$. Inception Score is the KL divergence of probability distribution of $p(y|x_i)$ and $p(y)$. The equation is given by

$$IS = exp\left(\frac{1}{X}\sum_{x_i \in X} p(y|x_i) \log \frac{p(y|x_i)}{p(y)}\right). \quad (6)$$

The larger of the value difference between $p(y|x_i)$ and $p(y)$ is, the larger Inception Score is.

VI. CAPSULE GAN ARCHITECTURE

We propose two architectures of Capsule GAN. One is to use the layer of the discriminator as the input to the generator (Capsule GAN2). The other is to use Capsule Network in generator such as CNNs (Capsule GAN3). The discriminator architectures are the same.

A. Capsule GAN2 (Using DIgitCaps Layer in Generator))

In this architecture, Capsule Network is used as is in the discriminator. The structure is shown in fig.4. The inputs to the discriminator are training images (real) and generated images by the generator (fake). The discriminator outputs whether the input image is real or fake. DigitCaps layer of Capsule Network holds the features of images. Therefore, the outputs of DigitCaps layer are the features of the real images and the fake images. The generator uses DigitCaps layer from the discriminator. DigitCaps layer is multiplied by latent variables. They are regarded as the input of the generator. The generator uses the outputs of DigitCaps layer, which holds the feature of images. We use only DigitCaps layer of real images. We found that the quality of the generated images is poor when the generator uses the feature of the fake images. This is because the generator uses the feature of fake images early in training, despite the feature used is of low quality. The experimental result of this phenomena is shown in Section VII-B. After receiving the outputs of DigitCaps layer as inputs, the generator calculates deconvolution such as DCGAN, and outputs images.

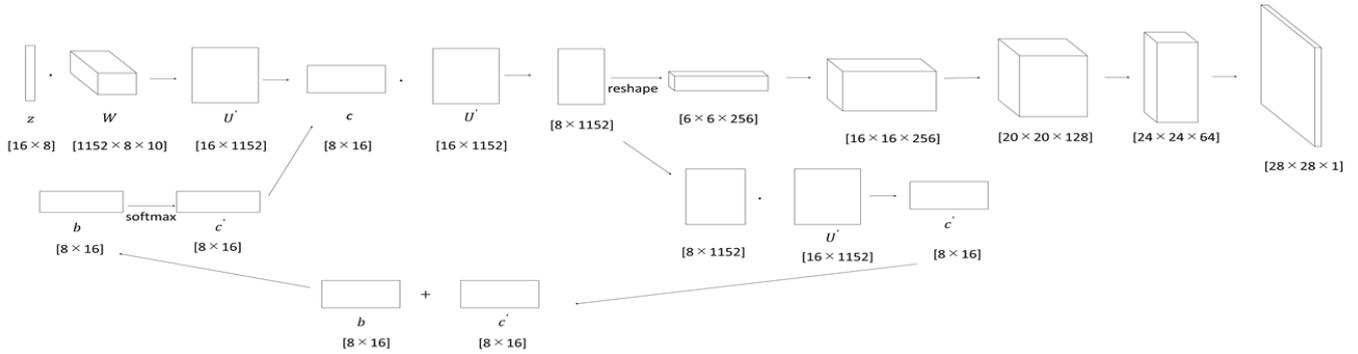

Fig.5. The generator architecture of Capsule GAN3. It is the reverse operation of recognition process in Capsule Network. This architecture is for generating 28×28 pixel images, MNIST. Before deconvolution layers, input values are calculated in dynamic routing.

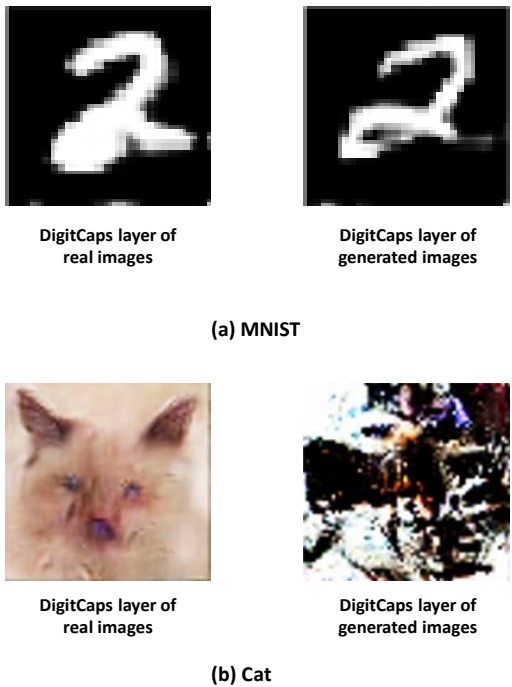

(a) MNIST

(b) Cat

Fig.6. Generated images by Capsule GAN2.

In this approach, Capsule GAN uses Capsule Network in both the discriminator and the generator by using DigitCaps layer.

*B. Capsule GAN3 (Using Capsule Network in Generator)*

In this architecture, Capsule GAN uses Capsule Network for the generator as well as the discriminator. The discriminator architecture is the same as that introduced in Capsule GAN2. The generator architecture is shown in Fig.5. It is the reverse operation of recognition process in Capsule Network. The values needed for generating images are increased by dynamic routing before the deconvolution layers. It stabilizes the quality of the generated images. Figure 5 shows the case when the generator produces 28×28 pixel images. When using Capsule Network in the generator, dynamic routing parameters have a significant effect on the output images. We have not adjust the parameters of generating 64×64 pixel color images. Therefore, we performed an experiment using only MNIST and Fashion-MNIST on this architecture.

Capsule Network is incorporated into the discriminator and the generator.

## VII. EXPERIMENT

We evaluate the performance of DCGAN, Capsule GAN by Gadirov et al. [5] (Capsule GAN1), Capsule GAN2 and Capsule GAN3. In addition, we performed an experience of the architecture of Capsule GAN 2 proposed in Section V-A.

*A. Dataset*

We use three datasets in the experiments, MNIST, Fashion-MNIST and "cat images". These datasets are as follows:

TABLE I. DCGAN ARCHITECTURE

| Discriminator | | Generator | |
|---|---|---|---|
| input | Output | input | Output |
| Conv, LeakyRelu | $32 \times 14 \times 14$ | Dense | 6272 |
| Dropout | $32 \times 14 \times 14$ | Reshape | $128 \times 7 \times 7$ |
| Conv, BN, LeakyRelu | $64 \times 8 \times 8$ | Deconvolution, BN, Relu | $128 \times 14 \times 14$ |
| Dropout | $64 \times 8 \times 8$ | Deconvolution, BN, Relu | $64 \times 28 \times 28$ |
| Conv, BN, LeakyRelu | $128 \times 4 \times 4$ | Conv, Tanh | $1 \times 28 \times 28$ |
| Dropout | $128 \times 4 \times 4$ | – | – |
| Conv, BN, LeakyRelu | $256 \times 4 \times 4$ | – | – |
| Dropout | $256 \times 4 \times 4$ | – | – |
| Dense | 1 | – | – |

TABLE II. CAPSULE GAN2 ARCHITECTURE

| Discriminator | | Generator | |
|---|---|---|---|
| input | Output | input | Output |
| Conv, LeakyRelu | $256 \times 20 \times 20$ | Multiply, BN, LeakyRelu | $16 \times 100$ |
| Primary, squash | $256 \times 6 \times 6$ | Weight, BN, LeakyRelu | 100 |
| DigitCap | $16 \times 10$ | Dense | 6272 |
| Mask | 16 | Reshape | $128 \times 7 \times 7$ |
| Dense | 1 | Deconvolution, BN, Relu | $128 \times 14 \times 14$ |
| – | – | Deconvolution, BN, Relu | $64 \times 28 \times 28$ |
| – | – | Conv, Tanh | $1 \times 28 \times 28$ |

TABLE III. CAPSULE GAN3 ARCHITECTURE

| Discriminator | | Generator | |
|---|---|---|---|
| Input | Output | input | Output |
| Conv, LeakyRelu | $256 \times 20 \times 20$ | Reshape | $8 \times 16$ |
| Primary, squash | $256 \times 6 \times 6$ | DigitCaps | $8 \times 1152$ |
| DigitCap | $16 \times 10$ | Reshape | $256 \times 6 \times 6$ |
| Mask | 16 | Deconvolution, BN, Relu | $256 \times 16 \times 16$ |
| Dense | 1 | Deconvolution, BN, Relu | $128 \times 20 \times 20$ |
| – | – | Deconvolution, BN, Relu | $64 \times 24 \times 24$ |
| – | – | Deconvolution, Tanh | $1 \times 28 \times 28$ |

*1) MNIST* [8]*:* MNIST is a dataset consisting of images of handwritten digits from 0 to 9 and digits labels. MNIST contains 60000 images for training and 10000 images for verfication. These images are $28 \times 28$ pixel grayscale images. We use the 60000 training images in the experiment.

*2) Fashion-MNIST* [9]*:* Fashion-MNIST is a dataset consisting of images of clothes such as shirts and pants and their labels. Like MNIST, it contains 60000 images for training and 10000 images for verfication. These images are $28 \times 28$ pixel grayscale images. We use the 60000 training images in the experiment.

*3) Cat Images* [10]*:* We use 4978 cat images from "The Oxford- IIIT Pet Dataset" [21] publised by Oxford University. In addition, we crawled cat images to increase the number of images, and a total of 7896 cat images were prepared. These images are $64 \times 64$ pixel color images.

### B. Experiment of the Architecture

First, we performed an experiment on the architecture shown in Fig.1 (c). In this architecture, the generator uses DigitCaps layer from the discriminator as inputs. The inputs of the discriminator are the training images and the images generated by the generator during training. Therefore, DigitCaps layer has two types of outputs, with the feature of the training images and the generated images. As a result, there are also two types of the DigitCaps layer used as generator inputs during training. The MNIST and cat images generated using each DigitCaps layer are shown in Fig.6. We can see that the generated MNIST images do not change much in Fig.6. However, the quality of cat images using DigitCaps layer of the generated images is quite different from the one of the real images. It is due to difference in the complexity of the generated images. In GAN training, the discriminator learning tends to proceed ahead of the generator learning [18]. Thus, the discriminator learning proceeded before the generator outputs high quality cat images due to the complexity of the images. Training converged before the discriminator and the generator learning was balanced. Consequently, the generator failed to generate high quality images with DigitCaps layer of the generated images.

For this reason, we use only DigitCaps layer of the training images in the generator.

### C. Comparative Experiment

We performed a comparative experiment on DCGAN, Capsule GAN1, Capsule GAN2 and Capsule GAN3. Capsule GAN1 uses Capsule Network in the discriminator only. Capsule GAN2 and Capsule GAN3 use Capsule Network in both the discriminator and the generator. The architecture of DCGAN, Capsule GAN2 and Capsule GAN3 are shown in Table I, Table II and Table III. These architectures are for generating $28 \times 28$ pixel images. For Capsule GAN1, the discriminator architecture is the same as Table II and the generator architecture is the same as Table I. We compared the images generated by the four GANs above for MNIST and Fashion-MNIST. All hyperparameters are the same. We use Adam [22] as the activation function. These GANs were trained until convergence. For cat images, we compared the images generated by DCGAN, Capsule GAN1 and Capsule GAN2. We performed an experiment on Capsule GAN3, but the parameters did not converge because of very sensitive setting of parameters shown as N/A in Table IV.

TABLE IV. TABLE STYLES

| | Images | DCGAN | Capsule GAN (Only Discriminator) | Capsule GAN (Using DisitCaps Layer) | Capsule GAN |
|---|---|---|---|---|---|
| **Inception Score** | MNIST | 2.32 | 2.35 | 2.37 | **2.57** |
| | FahionMNIST | 4.39 | 4.34 | 4.48 | **4.54** |
| | Cat | 3.86 | 4.37 | **4.59** | N/A |

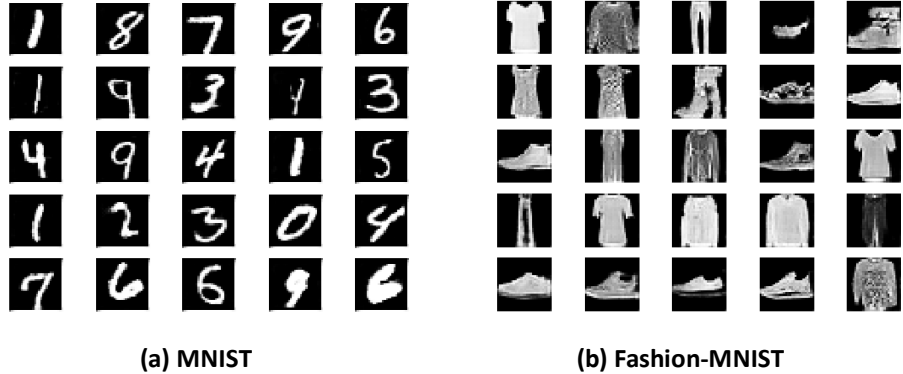

(a) MNIST  (b) Fashion-MNIST

Fig.7. Generated images by Capsule GAN3

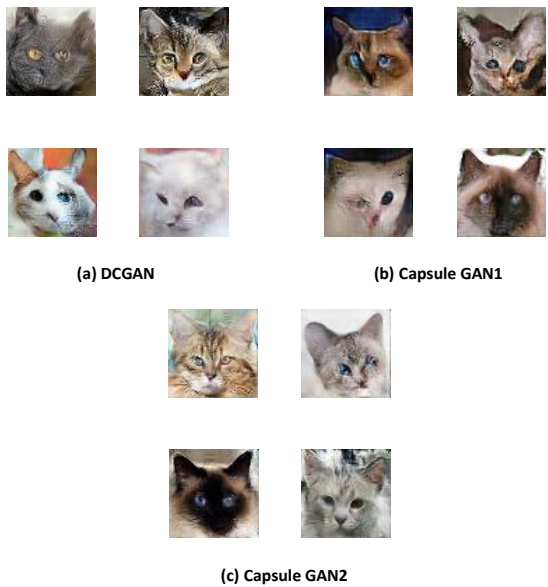

Fig.8. Generated cat images

Table IV shows the Inception Score of the images generated by each GAN. Fig.7 shows the images generated by Capsule GAN3. The Inception Score of Capsule GAN3 is largest for MNIST and Fashion-MNIST. The second is Capsule GAN2. According to these results, GAN outputs better images by using Capsule Network in both discriminator and generator. Fig.8 shows the cat images generated by each GAN. From Table IV, the Inception Score of Capsule GAN2 is the largest in cat images. We can also see that Capsule GAN2 generates higher quality images in Fig.8. The number of cat images is smaller than that of MNIST and Fashion-MNIST. However, Capsule GAN2 generates images of some quality.

The results show that GANs using Capsule Network in both the discriminator and the generator generates higher quality images than GANs using CNN. In generating MNIST and Fashion-MNIST images, Capsule GAN3 generates highest quality images. It indicates that Capsule GAN3 generates better color images or large-sized images by adjusting the dynamic routing parameters of the generator.

VIII. CONCLUSION

This paper proposes Capsule GAN using Capsule Network in both the discriminator and the generator. We introduce two architectures. One is to use DigitCaps layer of the discriminator as the input to the generator. The other is to use Capsule Network in the generator such as CNNs. Both discriminators use Capsule Network as it is. We compared the performance of Capsule GANs proposed in this paper with DCGAN using CNNs and Capsule GAN using Capsule Network in only the discriminator. Consequently, we found that Capsule GAN proposed in this paper generated higher quality images. From this reason, it indicates that using Capsule Network for both the discriminator and the generator of GAN is effective.

We plan to adjust the parameters of the generator which uses the reverse operation of recognition process in Capsule Network in the future. If a large or color image can be handled, it will be possible to generate a high-quality image than that of

GAN using CNNs. The architecture we propose is the basic part of GANs. Therefore, it is possible to adopt many previously proposed methods of improving GANs quality and stabilizing training. We will improve the performance of Capsule GAN by adopting various methods.